\documentclass[10pt, conference, compsocconf]{IEEEtran}

\newif\ifapx

\newcommand{\ourmaintitle}{Telling Cause from Effect\\ using MDL-based Local and Global Regression}

\newcommand{\ourmethod}{\textsc{Slope}\xspace}
\newcommand{\oururl}{\url{http://eda.mmci.uni-saarland.de/slope/}}
\newcommand{\codeurl}{\oururl}

\usepackage{pgfplots}
\usepgfplotslibrary{fillbetween}
\usepackage{latexsym}
\usepackage{subcaption}
\usepackage{balance}
\usepackage[ruled, vlined, nofillcomment, linesnumbered]{algorithm2e}
\usepackage[noenumitem]{eda} 

\usepackage[pdftitle={Slope}, pdfauthor={Alexander Marx and Jilles Vreeken}, hidelinks, hyperfootnotes = false]{hyperref}

\graphicspath{ {./} }

\ifpdf
\usetikzlibrary{external}
\fi

\SetKwComment{tcpas}{\{}{\}}
\SetCommentSty{textnormal}
\SetArgSty{textnormal}
\SetKw{False}{false}
\SetKw{True}{true}
\SetKw{Null}{null}
\SetKwInOut{Output}{output}
\SetKwInOut{Input}{input}
\SetKw{AND}{and}
\SetKw{OR}{or}
\SetKw{Continue}{continue}

\newcommand{\cure}{\textsc{Cure}\xspace}
\newcommand{\resit}{\textsc{Resit}\xspace}

\newcommand{\anm}{\textsc{ANM}\xspace}
\newcommand{\igci}{\textsc{IGCI}\xspace}
\newcommand{\ourmethodD}{\textsc{SlopeD}\xspace}

\newcommand{\fitF}{\textsc{FitDeterministic}\xspace}

\newcommand{\XtoY}{{X \rightarrow Y}\xspace}
\newcommand{\YtoX}{{Y \rightarrow X}\xspace}

\newcommand{\DXY}{\Delta_{\XtoY}\xspace}
\newcommand{\DYX}{\Delta_{\YtoX}\xspace}

\newcommand{\DhXY}{\hat{\Delta}_{\XtoY}\xspace}
\newcommand{\DhYX}{\hat{\Delta}_{\YtoX}\xspace}

\newcommand{\fClass}{\mathcal{F}\xspace}
\newcommand{\Params}{\Phi\xspace}
\newcommand{\param}{\phi\xspace}

\newcommand{\res}{\tau\xspace}
\newcommand{\conf}{\mathbb{C}\xspace}

\newcommand\independent{\protect\mathpalette{\protect\independenT}{\perp}}
\def\independenT#1#2{\mathrel{\rlap{$#1#2$}\mkern2mu{#1#2}}}

	\tikzstyle{flatlabel}  = [above, font = \tiny, inner sep = 1pt, text = black]
	\tikzstyle{flatlabelb}  = [below, font = \tiny, inner sep = 1pt, text = black]
	\tikzstyle{slopelabel}  = [sloped, above, font = \tiny, inner sep = 1pt, text = black]
	\tikzstyle{slopelabelb}  = [sloped, below, font = \tiny, inner sep = 1pt, text = black]

\usepackage{tikz}
\usepackage{pgfplots}
\usepackage{pgfplotstable}
\usetikzlibrary{arrows,shapes,positioning,fit,calc,matrix,trees}
\usetikzlibrary{external}

\pgfdeclareplotmark{btriangle}{%
  \pgfpathmoveto{\pgfpoint{-\pgfplotmarksize}{\pgfplotmarksize}}
  \pgfpathlineto{\pgfpoint{0pt}{-\pgfplotmarksize}}
  \pgfpathlineto{\pgfpoint{\pgfplotmarksize}{\pgfplotmarksize}}
  \pgfpathlineto{\pgfpoint{-\pgfplotmarksize}{\pgfplotmarksize}}
  \pgfusepathqstroke
}
\pgfdeclareplotmark{btriangle*}{%
  \pgfpathmoveto{\pgfqpoint{0pt}{-\pgfplotmarksize}}
  \pgfpathlineto{\pgfqpointpolar{150}{\pgfplotmarksize}}
  \pgfpathlineto{\pgfqpointpolar{30}{\pgfplotmarksize}}
  \pgfpathclose
  \pgfusepathqfillstroke
}

\pgfdeclareplotmark{ltriangle}{%
  \pgfpathmoveto{\pgfqpoint{\pgfplotmarksize}{0pt}}
  \pgfpathlineto{\pgfqpointpolar{-120}{\pgfplotmarksize}}
  \pgfpathlineto{\pgfqpointpolar{120}{\pgfplotmarksize}}
  \pgfpathclose
  \pgfusepathqstroke
}
\pgfdeclareplotmark{ltriangle*}{%
  \pgfpathmoveto{\pgfqpoint{\pgfplotmarksize}{0pt}}
  \pgfpathlineto{\pgfqpointpolar{-120}{\pgfplotmarksize}}
  \pgfpathlineto{\pgfqpointpolar{120}{\pgfplotmarksize}}
  \pgfpathclose
  \pgfusepathqfillstroke
}

\pgfdeclareplotmark{rtriangle}{%
  \pgfpathmoveto{\pgfqpoint{-\pgfplotmarksize}{0pt}}
  \pgfpathlineto{\pgfqpointpolar{-60}{\pgfplotmarksize}}
  \pgfpathlineto{\pgfqpointpolar{60}{\pgfplotmarksize}}
  \pgfpathclose
  \pgfusepathqstroke
}

\definecolor{yafaxiscolor}{rgb}{0.3, 0.3, 0.3}
\definecolor{yafcolor1}{rgb}{0.4, 0.165, 0.553}
\definecolor{yafcolor2}{rgb}{0.949, 0.482, 0.216}
\definecolor{yafcolor3}{rgb}{0.47, 0.549, 0.306}
\definecolor{yafcolor4}{rgb}{0.925, 0.165, 0.224}
\definecolor{yafcolor5}{rgb}{0.141, 0.345, 0.643}
\definecolor{yafcolor6}{rgb}{0.965, 0.633, 0.267}
\definecolor{yafcolor7}{rgb}{0.627, 0.118, 0.165}
\definecolor{yafcolor8}{rgb}{0.878, 0.475, 0.686}

\pgfplotscreateplotcyclelist{yaf}{%
{yafcolor1,mark options={scale=0.45},mark=*},
{yafcolor5,mark options={scale=0.60},mark=triangle*},
{yafcolor3,mark options={scale=0.45},mark=square*},
{yafcolor4,mark options={scale=0.70},mark=btriangle*},
{yafcolor8,mark options={scale=0.60},mark=ltriangle*},
{yafcolor7,mark options={scale=0.60},mark=rtriangle*},
{yafcolor2,mark options={scale=0.60},mark=diamond*},
{yafcolor6,mark options={scale=0.60},mark=pentagon*}}

\tikzset{
precise pin/.style args={[#1][#2]#3:#4}{
    pin={[inner sep=0pt, #1, label={[append after command={
		node [#2,
			outer sep = 0pt,
			inner sep=0pt,
			at=(\tikzlastnode),
			anchor=#3+180 ] {#4} } ]center:{}}]#3:{}}
}}

\pgfplotsset{
	clip = false,
	clip marker paths = true,
	tick align=outside,
	x tick label style = {font=\scriptsize, yshift = 1pt},
	y tick label style = {font=\scriptsize, xshift = 1pt},
	major tick length = 2pt,
    every axis y label/.style = {at = {(ticklabel cs:0.5)}, rotate=90, anchor=center, font=\scriptsize, xshift = 2pt},
	every axis x label/.style = {at = {(ticklabel cs:0.5)}, anchor=center, font=\scriptsize, yshift = -2pt},
	axis y line*=left, axis x line*=bottom,
        enlargelimits = 0.03
}
\tikzstyle{every pin}=[font=\footnotesize, inner sep = 0pt, distance=2em]
\tikzstyle{every pin edge}=[line width = 0.1pt, pin distance = 2em]

\newlength{\myheight}
\newlength{\mywidth}

\newcommand{\legDist}{Distance ($\savg{\distc{}}$)}
\newcommand{\legJacc}{Jacc.\ dist.\ ($\savg{\jacc{}}$)}
\newcommand{\legCover}{Coverage ($\cover{}$)}
\newcommand{\legDens}{Density ($\savg{\density{}}$)}

\colorlet{graphcl1}{yafcolor1!50}
\colorlet{graphcl2}{yafcolor4!30}
\colorlet{graphcl3}{yafcolor2!50}
\colorlet{graphcl4}{yafcolor5}
\colorlet{graphcl5}{yafcolor4}
\colorlet{graphcl6}{yafcolor6}

\tikzstyle{graphedge} = [black, thick, opacity = 0.5]
\tikzstyle{graphnode} = [draw = black, circle, line width = 0pt, text = black, inner sep = 0.5pt, text width = 10pt, align = center]
\tikzstyle{outliernode} = [circle, line width = 0pt, draw, text = black, fill = white, inner sep = 0.5pt, text width = 10pt, align = center]

\tikzstyle{toyedge} = [->, black, thick, bend left = 10, yafcolor5]
\tikzstyle{toynode} = [draw = black, thick, circle, line width = 0pt, text = black, inner sep = 0pt, text width = 13pt, align = center]
\tikzstyle{groupline} = [black, thick, dashed]

\makeatletter
\tikzset{multicircle/.style  args={#1, (#2)}{%
 alias=tmp@name, %
  postaction={%
    insert path={
     \pgfextra{%
     \pgfpointdiff{\pgfpointanchor{\pgf@node@name}{center}}%
                  {\pgfpointanchor{\pgf@node@name}{east}}%
     \pgfmathsetmacro\insiderad{\pgf@x}%
     \foreach \c [count=\ci from = 0, evaluate=\ci as \angle using 360 - (\ci) * #1] in {#2}%
        \fill[\c] (\pgf@node@name.center)  -- ++(0:\insiderad-\pgflinewidth) arc (0:\angle:\insiderad-\pgflinewidth)--cycle;%
        }}}}}
\makeatother

\pgfplotsset{
  boxplot/box width/.initial=1em,
  solid boxes/.style={
    mark=x,
    boxplot/draw direction=y,
    boxplot/whisker extend=0,
    boxplot/draw/median/.code={%
      \draw[mark size=2pt,/pgfplots/boxplot/every median/.try]
        \pgfextra
        \pgftransformshift{
          \pgfplotsboxplotpointabbox
            {\pgfplotsboxplotvalue{median}}
            {0.5}
        }
        \pgfsetfillcolor{white}
        \pgfuseplotmark{*}
        \endpgfextra
      ;
    },
    boxplot/draw/box/.code={
      \draw[fill,/pgfplots/boxplot/every box/.try]
        ($(boxplot box cs:\pgfplotsboxplotvalue{lower quartile},0.5)!0.5\pgfkeysvalueof{/pgfplots/boxplot/box width}!(boxplot box cs:\pgfplotsboxplotvalue{lower quartile},0)$)
        rectangle
        ($(boxplot box cs:\pgfplotsboxplotvalue{upper quartile},0.5)!0.5\pgfkeysvalueof{/pgfplots/boxplot/box width}!(boxplot box cs:\pgfplotsboxplotvalue{upper quartile},1)$)
      ;
    }
  },
}

\begin{document}
\setlength{\pdfpagewidth}{8.5in}
\setlength{\pdfpageheight}{11in}

\title{\ourmaintitle}

\author{\IEEEauthorblockN{Alexander Marx and Jilles Vreeken}
\IEEEauthorblockA{Max Planck Institute for Informatics and Saarland University\\
Saarland Informatics Campus, Saarbr\"{u}cken, Germany\\
\{amarx, jilles\}@mpi-inf.mpg.de}
}

\date{}

\maketitle
   
\begin{abstract}
We consider the fundamental problem of inferring the causal direction between two univariate numeric random variables $X$ and $Y$ from observational data. The two-variable case is especially difficult to solve since it is not possible to use standard conditional independence tests between the variables.

To tackle this problem, we follow an information theoretic approach based on Kolmogorov complexity and use the Minimum Description Length (MDL) principle to provide a practical solution. In particular, we propose a compression scheme to encode local and global functional relations using MDL-based regression. We infer $X$ causes $Y$ in case it is shorter to describe $Y$ as a function of $X$ than the inverse direction.
In addition, we introduce \ourmethod, an efficient linear-time algorithm that through thorough empirical evaluation on both synthetic and real world data we show outperforms the state of the art by a wide margin.
\end{abstract} 

\section{Introduction}
\label{sec:intro}

Telling cause from effect from observational data is one of the fundamental problems in science~\cite{spirtes:00:book,pearl:09:book}. We consider the problem of inferring the most likely direction between two univariate numeric random variables $X$ and $Y$. That is, we are interested in identifying whether $X$ causes $Y$, whether $Y$ causes $X$, or whether they are merely correlated. 

Traditional methods, that rely on conditional independence tests, cannot decide between the Markov equivalent classes of $\XtoY$ and $\YtoX$~\cite{pearl:09:book}. Recently, it has been postulated however that if $\XtoY$, there exists an independence between the marginal distribution of the cause, $P(X)$, and the conditional distribution of the effect given the cause, $P(Y\mid X)$~\cite{shimizu:06:anm,janzing:10:algomarkov}. The state of the art exploits this asymmetry in various ways, and overall obtain up to $70\%$ accuracy on a well-known benchmark of cause-effect pairs~\cite{sgouritsa:15:cure,hoyer:09:nonlinear,peters:14:continuousanm,janzing:12:igci,mooij:16:pairs}. In this paper we break this barrier, and give an elegant score that is computable in linear-time and obtains over $82\%$ accuracy on the same benchmark. 

\begin{figure}[t]
	\centering
	\includegraphics[]{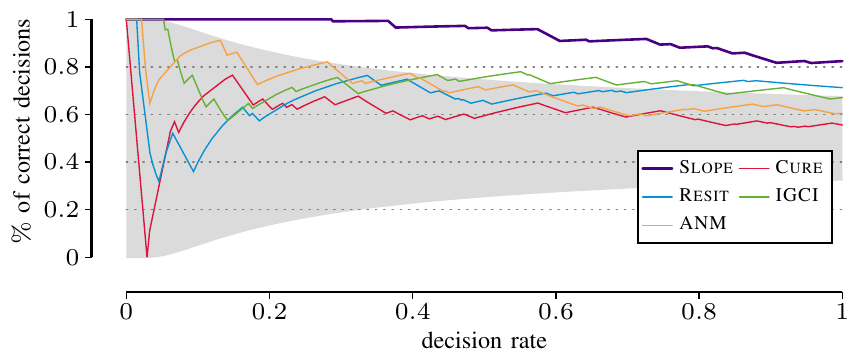}
	\caption{[Higher is better] 
	Weighted accuracy of our method, \ourmethod, versus the state of the art in causal inference for univariate numeric pairs as identified in a recent survey~\cite{mooij:16:pairs}, i.e., \cure~\cite{sgouritsa:15:cure}, \resit~\cite{peters:14:continuousanm}, \igci~\cite{janzing:12:igci} and ANM-pHSIC~\cite{hoyer:09:nonlinear} on the Tuebingen benchmark data set (98 pairs). The gray area indicates the $95\%$ confidence interval of a fair coin toss.}
	\label{fig:decision_rate_intro}
\end{figure}

To illustrate its strength, we show the results of our method, called \ourmethod, and four state-of-the-art methods over this benchmark in Fig.~\ref{fig:decision_rate_intro}. In particular, we show the accuracy over the top-$k$ pairs ranked according to the score at hand, the so-called decision rate. The plot shows that \ourmethod leads by large margin over its competitors; it is $100\%$ accurate over the 32 pairs it is most certain about, and is $90\%$ accurate over its top-74 out of 98 pairs. Unlike its competitors, its accuracies are strongly significant with regard to the $95\%$ confidence interval of a fair coin flip. Moreover, our score comes with a natural significance test that allows us to weed out insignificant results; when we consider the $83$ significant pairs only we find that \ourmethod is even $85\%$ accurate.

We base our method on the algorithmic Markov condition, a recent postulate by Janzing and Sch\"{o}lkopf~\cite{janzing:10:algomarkov}, that states that if $X$ causes $Y$, the factorization of the joint distribution $P(X,Y)$ in the causal direction has a simpler description---in terms of Kolmogorov complexity---than that in the anti-causal direction. That is, if $\XtoY$, $K(P(X))+K(P(Y\mid X)) \leq K(P(Y))+K(P(X\mid Y))$. As any physical process can be modelled by a Turing machine, this ideal score can detect any causal dependence that can be explained by a physical process. However, Kolmogorov complexity is not computable, so we need a practical instantiation of this ideal. In this paper we do so using the Minimum Description Length (MDL) principle, which provides a statistically well-founded approximation of Kolmogorov complexity. 

\begin{figure}
	\centering
       \begin{minipage}[t]{.5\linewidth}
       \includegraphics[]{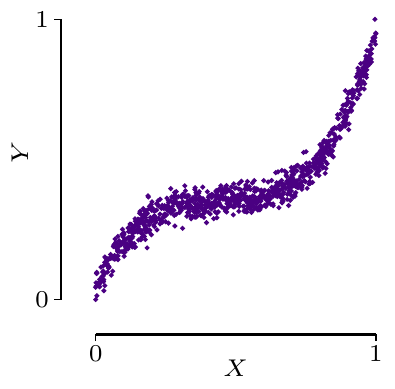}
	\subcaption{deterministic}\label{fig:deterministic_sample}
   	\end{minipage}%
   	\begin{minipage}[t]{.5\linewidth}
	\includegraphics[]{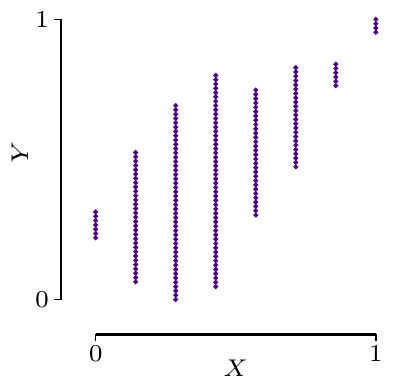}
	\subcaption{non-deterministic}\label{fig:non-deterministic_sample}
    \end{minipage}%
	\caption{Example deterministic and non-deterministic data. In both cases the ground truth is $X$ causes $Y$. The left-hand data is generated using a quadratic function with Gaussian noise, whereas the right-hand data is generated using a non-deterministic function.}
	\label{fig:scatter}
\end{figure}

Simply put, we propose to fit a regression model from $X$ to $Y$, and vice versa, measuring both the complexity of the function, as well as the error it makes in bits, and infer that causal direction by which we can describe the data most succinctly. There is a little bit more to it, of course. We carefully construct an MDL score such that we can meaningfully compare between different types of functional dependencies, including linear, quadratic, cubic, reciprocal, and exponential functions. This way, for example, we will find that we can more succinctly describe the data in Fig.~\ref{fig:deterministic_sample} by a cubic function than with a linear function, as while it takes fewer bits to describe the latter function, it will take many more bits to describe the large error it makes. 

We do not only consider models that try to explain all the data with a single, global, deterministic regression function, but also allow for non-deterministic models. That is, we consider compound regression functions that in addition to a global deterministic function additionally include regression functions for local parts of the data corresponding to specific, duplicated $X$ values. For example, consider the data in Fig.~\ref{fig:non-deterministic_sample}, where the $Y$ values belonging to a single $X$ value clearly show more structure than the general linear trend. In contrast, if we rotate the plot by $90$ degrees, we do not observe the same regularities for the $X$ values mapped to a single $Y$ value. In many cases, e.g., $Y=1$ there is only one mapping $X$ value. We can exploit this asymmetry by considering local regression functions per value of $X$, each individually fitted but as we assume all noise to be of the same type, all should be of the same function class. In this particular example, we therewith correctly infer that $X$ causes $Y$. The MDL principle prevents us from overfitting, as such local functions are only included if they aid global compression. Last, but not least, we give a linear-time algorithm, \ourmethod, to compute this score.

As we model $Y$ as a function of $X$ and noise, our approach is somewhat reminiscent to causal inference based on Additive Noise Models (ANMs)~\cite{shimizu:06:anm}, where one assumes that $Y$ is generated as a function of $X$ plus additive noise, $Y = f(X) + N$ with $X \independent N$. In the ANM approach, we infer $\XtoY$ if we can find a function from $X$ to $Y$ that admits an ANM, but cannot do so in the opposite direction. In practice, ANM methods often measure the independence between the presumed cause and the noise in terms of p-values, and infer the direction of the lowest p-value. As we will see, this leads to unreliable confidence scores---not the least because p-values are often strongly influenced by sample size~\cite{anderson:00:p-values}, but also as that a lower p-value does not necessarily mean that $H_1$ is more true, just that $H_0$ is very probably not true~\cite{anderson:00:p-values}. We will show that our score, on the other hand, is robust against sample size, and correlates strongly with accuracy. Moreover, it admits an elegant and effective analytical statistical test on the \emph{difference} in score between the two causal directions based on the no-hypercompressibility inequality~\cite{bloem:17:arxiv,grunwald:07:book}.

Our key contributions can be summarised as follows, we 
\begin{itemize}[noitemsep,topsep=0pt]
	\item[(a)] propose an MDL score for causal inference on pairs of univariate numeric random variables, 
	\item[(b)] formulate an analytic significance test based on compression,
	\item[(c)] show to model unobserved mechanisms via compound deterministic and non-deterministic functions, 
	\item[(d)] introduce the linear-time \ourmethod algorithm, 
	\item[(e)] give extensive empirical evaluation on synthetic and real-world data, including a case study
	\item[(f)] make all code, data generators, and data available.
\end{itemize}

The remainder of this paper is organised as usual. We first give a brief primer to Kolmogorov complexity and the Minimum Description Length principle in Sec.~\ref{sec:prelim}. In Sec.~\ref{sec:theory} we introduce our score based on the algorithmic independence of conditional, as well as a practical instantiation based on the MDL principle. We introduce the linear-time \ourmethod algorithm to efficiently compute the conditional score in Sec.~\ref{sec:algo}. Sec.~\ref{sec:related} discusses related work. We empirically evaluate \ourmethod in Sec.~\ref{sec:exps} and discuss the results in Sec.~\ref{sec:discussion}. We round up with conclusions in Sec.~\ref{sec:conclusion}.

\section{Preliminaries}
\label{sec:prelim}

In causal inference, the goal is to determine for two random variables $X$ and $Y$ that are statistically dependent whether it is more likely that $X$ \textit{causes} $Y$, denoted by $\XtoY$, or whether it is more likely that $Y$ causes $X$, $\YtoX$. In this paper we consider the case where $X$ and $Y$ are univariate and numeric. We work under the common assumption of causal sufficiency~\cite{budhathoki:16:origo,mooij:16:pairs,peters:14:continuousanm, verma:90:markov-equiv}. That is, we assume there is no hidden confounder variable $Z$ that causes both $X$ and $Y$.

We base our causal inference score on the notion of Kolmogorov complexity, which we will approximate via MDL. Below we give brief primers to these two main concepts. 

\subsection{Kolmogorov Complexity}

The Kolmogorov complexity of a finite binary string $x$ is the length of the shortest binary program $p^*$ for a universal Turing machine $\mathcal{U}$ that outputs $x$ and then halts~\cite{kolmogorov:65:information, vitanyi:93:book}. Formally, 
\[
K(x) = \min \{ |p| \mid p \in \{0,1\}^*, \mathcal{U}(p) = x \} \; .
\]
Simply put, $p^*$ is the most succinct \emph{algorithmic} description of $x$, and therewith Kolmogorov complexity of $x$ is the length of its ultimate lossless compression. Conditional Kolmogorov complexity, $K(x \mid y) \leq K(x)$, is then the length of the shortest binary program $p^*$ that generates $x$, and halts, given $y$ as input. 

The Kolmogorov complexity of a probability distribution $P$, $K(P)$, is the length of the shortest program that outputs $P(x)$ to precision $q$ on input $\langle x, q \rangle$~\cite{vitanyi:93:book}. More formally, we have
\[
K(P) = \min \left \{ |p| : p \in \{0,1\}^*, |\mathcal{U}(\langle x, \langle q, p \rangle \rangle) - P(x) | \leq 1/q \right \} \; .
\]
The conditional, $K(P \mid Q)$, is defined similarly except that the universal Turing machine $\mathcal{U}$ now gets the additional information $Q$. The algorithmic mutual information between two distributions $P$ and $Q$ is $I(P : Q) = K(P) - K(P \mid Q^*)$, where $Q^*$ is the shortest binary program for $Q$. For more details on Kolmogorov complexity see~\cite{vitanyi:93:book}.

\subsection{Minimum Description Length Principle}
 
Kolmogorov complexity is, however, not computable~\cite{vitanyi:93:book}, but we can approximate it in a well-founded and computable way through the Minimum Description Length (MDL) principle~\cite{rissanen:78:mdl, grunwald:07:book}. Conceptually, instead of all programs, Ideal MDL considers only those for which we know that they output $x$ and halt, i.e., lossless compressors. Formally, given a model class $\mathcal{M}$, MDL identifies the best model $M \in \mathcal{M}$ for data $D$ as the one minimizing 
\[
L(D, M) = L(M) + L(D \mid M) \; ,
\]
where $L(M)$ is the length in bits of the description of $M$, and $L(D \mid M)$ is the length in bits of the description of data $D$ given $M$. This is known as two-part, or \emph{crude} MDL. There also exists one-part, or \emph{refined} MDL. Although refined MDL has theoretically appealing properties, it is only efficiently computable for a small number of model classes.

To use MDL in practice we need to define a model class, and how to encode a model, resp. the data given a model, into bits. Note that in MDL we are only concerned with optimal code \emph{lengths}, not actual codes---our goal is to measure the \emph{complexity} of a dataset under a model class, after all~\cite{grunwald:07:book}. 

\section{Information Theoretic Causal Inference}
\label{sec:theory}

In this section, we first introduce how to infer causal directions using Kolmogorov complexity. Thereupon, we show how to obtain a computable score based on the MDL principle. 

\subsection{Causal Inference by Kolmogorov Complexity}

A central postulate in causal inference concerns the algorithmic independence of conditionals. For multiple random variables, this postulate is defined as follows~\cite{janzing:10:algomarkov}. 

\vspace{0.5em}
\noindent\textbf{Algorithmic Independence of Conditionals:}
\textit{A causal hypothesis is only acceptable if the shortest description of the joint density $P$ is given by the concatenation of the shortest description of the Markov kernels. Formally, we write}
\begin{equation}
K(P(X_1,\dots,X_n)) \stackrel{+}{=} \sum_j K(P(X_j \mid PA_j)) \; , \label{eq:markov}
\end{equation}
\textit{which holds up to an additive constant independent of the input, and where $PA_j$ corresponds to the parents of $X_j$ in a causal directed acyclic graph (DAG).}
\vspace{0.5em}

As we consider two variables, $X$ and $Y$, either $X$ is the parent of $Y$ or the other way round. That is, either
\begin{align}
K(P(X,Y)) &\stackrel{+}{=} K(P(X)) + K(P(Y \mid X)) \; \text{, or} \\
K(P(X,Y)) &\stackrel{+}{=} K(P(Y)) + K(P(X \mid Y)) \; .
\end{align}
In other words, two valid ways to describe the joint distribution of $X$ and $Y$ include to first describe the marginal distribution $P(X)$ and then the conditional distribution $P(Y \mid X)$, or first to describe $P(Y)$ and then $P(X\mid Y)$.

Thereupon, Janzing and Sch{\"{o}}lkopf formulated the postulate for algorithmic independence of Markov kernels~\cite{janzing:10:algomarkov}.

\vspace{0.5em}
\noindent\textbf{Algorithmic Independence of Markov Kernels:}
\textit{If $\XtoY$, the marginal distribution of the cause $P(X)$ is algorithmically independent of the conditional distribution of the effect given the cause $P(Y \mid X)$, i.e., the algorithmic mutual information between the two will be zero,}
\begin{equation} \label{eq:janzing}
I(P(X):P(Y \mid X)) \stackrel{+}{=} 0 \; ,
\end{equation}
\textit{while this is not the case in the other direction.}
\vspace{0.5em}

Simply put, for the true causal direction, the marginal distribution of the cause is algorithmically independent of the conditional distribution of the effect given the cause. Building upon Eqs.~\eqref{eq:markov} and \eqref{eq:janzing}, Mooij et al.~\cite{mooij:10:mml} derived an inference rule stating that if $X$ causes $Y$, 
\begin{equation} \label{eq:mooij}
K(P(X)) + K(P(Y \mid X)) \le K(P(Y)) + K(P(X \mid Y))
\end{equation}
holds up to an additive constant. This means that if $\XtoY$, the description of the joint distribution $K(P(X,Y))$ of first describing the marginal distribution of the cause $K(P(X))$ and then describing the conditional distribution of the effect given the cause $K(P(Y \mid X))$, will be shorter than the other way around. 

Although Eq.~\eqref{eq:mooij} already allows for inferring the causal direction for a given pair, we obtain a more robust score, allowing for fair comparison of results independent of data sizes, when we normalise the result. In particular, Budhathoki and Vreeken~\cite{budhathoki:16:origo} recently proposed to normalise the scores with the sum of the description lengths for the marginal distributions. We therefore define our causal indicator as
\begin{equation}
\DXY = \frac{K(P(X)) + K(P(Y \mid X))}{K(P(X)) + K(P(Y))} \; , \label{eq:korigo}
\end{equation}
and $\DYX$ in the same manner. Consequently, we infer $\XtoY$, if $\DXY < \DYX$, and $\YtoX$, if $\DXY > \DYX$ and do not decide if $\DXY = \DYX$.

The confidence of our score is $\conf = |\DXY - \DYX|$. The higher, the more certain we are that the inferred causal direction is correct. To avoid confusion, we want to emphasise that $\conf$ has nothing to do with a confidence interval, but can be used to rank results of several tests. Below, after introducing our practical score, we will show how we can in addition define a practical statistical test.

\subsection{Causal Inference by MDL}

As Kolmogorov complexity is not computable, we will instantiate $\DXY$ and $\DYX$ using the Minimum Description Length principle~\cite{vitanyi:93:book,grunwald:07:book}. In practice this means we will estimate $\DXY$ as 
\begin{align}
\DhXY &= \frac{L(X) + L(Y \mid X)}{L(X) + L(Y)}
\end{align}
where $L(X)$ is the length in bits of the description of the marginal distribution of $X$, $L(Y)$ that of the marginal distribution of $Y$, and $L(Y \mid X)$ that of the conditional distribution of $Y$ given $X$. 
We define $\DhYX$ analogue to $\DhXY$, and we infer $\XtoY$, if $\DhXY < \DhYX$, $\YtoX$, if $\DhXY > \DhYX$ and do not decide if $\DhXY = \DhYX$ or below a user-defined threshold. Like above, confidence $\conf$ is simply the absolute difference between $\DhXY$ and $\DhYX$. 

Considering the difference between the encoded lengths is related to, but not the same as considering the ratio of the posteriors; we also include the complexity of the model, which helps against overfitting. Intuitively, if the functions we find for the two directions both explain the data equally well, we prefer that direction that explains it using the simplest function. 

This leaves us to explain how we encode the data, and, most importantly, how we encode $L(Y \mid X)$. 

\subsubsection*{Intuition of the Conditional Encoding}

The general idea is simple: we use regression to model the data of $Y$ given $X$. That is, we model $Y$ as a function $f$ of $X$ and independent noise $N$, i.e. $Y = f(X) + N$. We do so by fitting a regression function $f$ over $X$ and treating the error it makes as Gaussian distributed noise. Naturally, the better $f(X)$ fits $Y$, the fewer bits we will have to spend on encoding errors. The more parameters $f(X)$ has, however, the more bits we will have to spend on encoding these. This way, MDL naturally balances the complexity of the model to that of the data~\cite{grunwald:07:book}.
For example, while a linear function is more simple to describe than a cubic one, the latter will fit the data plotted in Fig.~\ref{fig:deterministic_sample} so much better that MDL decides it is the better choice. 

A key idea in our approach is to consider not only single global deterministic regression functions $f_g$, which works well for deterministic data, but to also consider non-deterministic, or compound functions as models. That is, we consider models that besides the global regression function $f_g$ may additionally consist of \emph{local} regression functions $f_l$ that model $Y$ for those values $x$ of $X$ that non-deterministically map to multiple values of $Y$. That is, per such value of $X$, we take the associated values of $Y$, sort these ascending, and uniformly re-distribute them on $X$ over a fixed interval. We now see how well, just for these re-distributed points, we can fit a local regression model $f_l$. This way, we will for example be able to much more succinctly describe the data in Fig.~\ref{fig:non-deterministic_sample} than with a single global deterministic regression function, as we can now exploit the structure that the values of $Y$ have given a value of $X$, namely, being approximately equally spaced. To avoid overfitting we use MDL, and only allow a local function for a value of $X$ into our model if it provides a gain in overall compression. Since we assume that for the true causal model the data in the local components follows the same pattern, we only allow models in which all local functions are of the same type, e.g., all are linear, all are quadratic, etc.

In the following paragraphs, we formalise these ideas and define our cost functions. All logarithms are to base 2, and we use the common convention that $0 \log 0 = 0$.

\subsubsection*{Complexity of the Marginals}
We start by defining the cost for the marginal distributions, $L(X)$ and $L(Y)$, which mostly serve to normalise our causal indicators $\DhXY$ and $\DhYX$. As we beforehand do not know how $X$ or $Y$ are distributed, and do not want to incur any undue bias, we encode both using a uniform prior with regard to the data resolution $\res$ of $X$ and $Y$. That is, we have $L(X) = - n \log \res$, where $\res$ is the resolution of the data of $X$. Note that resolution $\res$ can be different between $X$ and $Y$---we specify how we choose $\res$ in the next section. We define $L(Y)$ analogue.

\subsubsection*{Complexity of the Conditional Model}

Formally, we write $F$ for the set of regression functions, or model, we use to encode the data of $Y$ given $X$. A model $F$ consists of at least one global regression function $f_g \in \fClass$, and up to $\mathit{dom}(X)$ local regression functions $f_l \in \fClass$, associated with individual values of $X$. We write $F_l$ for the set of local regression functions $f_l \in F_l$, and require that all $f_l \in F_l$ are of the same type. The description length, or encoded size, of $F$ is 
\begin{align}
L(F) =& L_{\mathbb{N}}(|F|) + \log { {|X| - 1} \choose {|F_l| - 1}} + \\
& 2\log(|\fClass|) + L(f_g) + \sum_{f_l \in F_l} L(f_l) \; ,
\end{align}
where we first describe the number of local functions using $L_\mathbb{N}$, the MDL optimal encoding for integers $z \geq 1$~\cite{rissanen:83:integers}, then map each $f_l$ to its associated value of $X$, after which we use $\log |\fClass|$ bits to identify the type of the global regression function $f_g$, and whenever $F_l$ is non-empty also $\log |\fClass|$ bits to identify the type of the local regression functions $f_l$, finally, we encode the functions themselves. Knowing the type of a function, we only need to encode its parameters, and hence
\begin{equation}
L(f) = \sum_{\param \in \Params_{f}} L_{\mathbb{N}}(s) + L_{\mathbb{N}}(\lceil \param \cdot 10^s \rceil ) + 1 \; ,
\end{equation}
where we encode each parameter $\param$ up to a certain precision $p$. We shift $\param$ by the smallest integer number $s$ such that $\param \cdot 10^s \ge 10^p$, i.e. $p=3$ means that we consider three digits. What remains is that we need to encode the shift, the shifted digit and the sign.

\subsubsection*{Complexity of the Conditional Data}

Reconstructing the data of $Y$ given $f(X)$ corresponds to encoding the errors, or deviations, the model makes. Since we fit our regression functions by minimizing the sum of squared errors, which corresponds to maximizing the likelihood under a Gaussian, it is a natural choice to encode the errors using a Gaussian distribution with zero-mean. 

Since we have no assumption on the standard deviation, we use the empirical estimate $\hat{\sigma}$ to define the standard deviation of the Gaussian. By doing so, the encoded size of the error of $F(X)$ with respect to the data of $Y$ corresponds to  
\begin{align}
L(Y \mid F, X) = \sum_{f \in F} \left(  \frac{n_f}{2} \left( \frac{1}{\ln 2} + \log 2 \pi \hat{\sigma}^2  \right) - n_f \log \res \right) \; ,
\end{align}
where $n_f$ is the number of data points for which we use a specific function $f \in F$. Intuitively, this score is higher the less structure of the data is described by the model and increases proportionally to the sum of squared errors.

\subsubsection*{Complexity of the Conditional}

Having defined the data and model costs above, we can now proceed and define the total encoded size of the conditional distribution of $Y$ given $X$ as 
\begin{align}
L(Y \mid X) &= L(F) + L(Y \mid F, X) \; . \label{eq:lyx}
\end{align}
By MDL we are after that model $F$ that minimises Eq.~\eqref{eq:lyx}. After discussing a significance test for our score, we will present the \ourmethod algorithm to efficiently compute the conditional score in the next section. 

\subsubsection*{Significance by Hypercompression}

Ranking based on confidence works well in practice. Ideally, we would additionally like to know the significance of an inference. It turns out we can define an appropriate hypothesis test using the no-hypercompressibility inequality~\cite{bloem:17:arxiv,grunwald:07:book}. In a nutshell, under the hypothesis that the data was sampled from the null-model, the probability that any other model can compresses it $k$ bits better is $P_0(L_0(x)-L(x) \geq k) \leq 2^{-k}$. 

This means that if we assume the null complexity, $L_0$, to be the least-well compressed causal direction, we can evaluate the probability of gaining $k$ bits by instead using the most-well compressed direction. Formally, if we write $L(\XtoY)$ for $L(X) + L(Y\mid X)$, and vice-versa for $L(\YtoX)$, $L_0$ would be $\max\{L(\XtoY), L(\YtoX)\}$. The probability that the data can be compressed $k = |L(\XtoY) - L(\YtoX)|$ bits better in the opposite direction then simply is $2^{-k}$. 

In fact, we can construct a stronger test by assuming that the data is not causated, but merely correlated. That is, we assume \emph{both} directions are wrong; the one compresses too well, the other compresses too poor. Following, if we assume these two to be equal in terms of exceptionality, the null complexity is the mean between the complexities of the two causal directions, i.e., $L_0 = \min \{ L(\XtoY),L(\YtoX) \} + |L(\XtoY)- L(\YtoX)|/2$. The probability of the best-compressing direction is then $2^{-k}$ with $k = |L(\XtoY) - L(\YtoX)|/2$. We can now set a significance threshold $\alpha$ as usual, such as $\alpha = 0.001$, and use this to prune out those cases where the difference in compression between the two causal directions is insignificant. We will evaluate this procedure, in addition to our confidence score, in the experiments. 

\section{The \ourmethod Algorithm}
\label{sec:algo}

With the framework defined in the previous section, we can determine the most likely causal direction and the corresponding confidence value. In this section we present the \ourmethod algorithm to efficiently compute the causal indicators. To keep the computational complexity of the algorithm linear, we restrict ourselves to linear, quadratic, cubic, exponential and reciprocal functions---although at the cost of extra computation this class may be expanded arbitrarily. We start by introducing the subroutine of \ourmethod that computes the conditional complexity of $Y$ given $X$.

\subsection{Calculating the Conditional Scores}

Algorithm \ref{alg:slope} describes the subroutine to calculate the conditional costs $L(Y \mid X)$ or $L(X \mid Y)$. We start with fitting a global function $f_g$ for each function class $c \in \fClass$ and choose the one $f_g$ with the minimum sum of data and model costs (line~\ref{alg:slope:deterministic}). Next, we add $f_g$ to the model $F$ and store the total costs (\ref{alg:slope:model1}--\ref{alg:slope:costsD}). For purely deterministic functions, we are done. 

If $X$ includes duplicate values, however, we need to check whether fitting a non-deterministic model leads to a gain in compression. To this end we have to check for each value $x_i$ of $X$ that occurs at least twice, whether we can express the ascendingly ordered corresponding $Y$ values, $Y_i$, as a function $f_l$ of uniformly distributed data $X_i$ between $[-t,t]$, where $t$ is a user-determined scale parameter (lines~\ref{alg:slope:getY}--\ref{alg:slope:newCosts}). If the model costs of the new local function $f_l$ are higher than the gain on the data side, we do not add $f_l$ to our model (\ref{alg:slope:checkBetter}). As it is fair to assume that for truly non-deterministic data the generating model for each local component is the same, we hence restrict all local functions to be of the same model class $c \in \fClass$. As final result, we return the costs according to the model with the smallest total encoded size. In case of deterministic data, this will be the model containing only $f_g$.

\begin{algorithm}[tb!]
	\caption{\textsc{ConditionalCosts}$(Y, X)$}
	\label{alg:slope}
	\Input{random variables $Y$ and $X$}
	\Output{score $L(Y \mid X)$}
	$F = \text{empty model}$\;
	$f_g = \fitF(Y \sim X, \fClass)$\; \label{alg:slope:deterministic}
	$F = F \cup f_g$\;  \label{alg:slope:model1}
	$s = s_g = L(F) + L(Y \mid F, X)$\;  \label{alg:slope:costsD}
	$X_u = \{ x \in X \mid \text{count}(x) \geq 2 \}$\;  \label{alg:slope:uniqueX}
	\ForEach{$\fClass_c \in \fClass$}{
		$s_c = s_g, \; F_c = F$\;
		\ForEach{$x_i \in X_u$}{
			$Y_i = \{ y \in Y \mid y \text{ maps to } x_i\}$\;  \label{alg:slope:getY}
			$X_i = \text{norm}(1:|Y_i|, \min=-t, \max=t)$\;  \label{alg:slope:genericX}
			$f_l=  \fitF(Y_i \sim X_i, \fClass_c)$\;  \label{alg:slope:non-deterministic}
			$\hat{s} = L(F_c \cup f_l) + L(Y \mid F_c \cup f_l, X)$\; \label{alg:slope:newCosts}
			\lIf{$\hat{s} < s_c$}{$s_c = \hat{s}, \; F_c = F_c \cup f_l$} \label{alg:slope:checkBetter}
		}
		\lIf{$s_c < s$}{$s = s_c$}  \label{alg:slope:checkBetterFc}
	}
	\Return{$s$}\;
\end{algorithm}

\subsection{Causal Direction and Confidence}

In the previous paragraph, we described Algorithm \ref{alg:slope}, which is the main algorithmic part of \ourmethod. Before applying it, we first normalise $X$ and $Y$ to be from the same domain and then determine the data resolutions $\res_X$ and $\res_Y$ for $X$ and $Y$. To obtain the data resolution, we calculate the smallest difference between two instances of the corresponding random variable. Next, we apply Algorithm \ref{alg:slope} for both directions to obtain $L(Y \mid X)$ and $L(X \mid Y)$. Subsequently, we estimate the marginals $L(X)$ and $L(Y)$ based on their data resolutions. This we do by modelling both as a uniform prior with $L(X) = -n \log \res_X$ and $L(Y) = -n \log \res_Y$. In the last step, we compute $\DhXY$ and $\DhYX$ and report the causal direction as well as the corresponding confidence value $\conf$.

\subsection{Computational Complexity}

To assess the computational complexity, we have to consider the score calculation and the fitting of the functional relations. The model costs are computed in linear time according to the number of parameters, whereas the data costs need linear time with regard to the number of data points $n$. Since we here restrict ourselves to relatively simple functions, we can fit these in time linear to the number of data points. To determine the non-deterministic costs, in the worst case we perform $n/2$ times $|\fClass|$ fits over two data points, which is still linear. In total the runtime complexity of \ourmethod hence is $O(n |\fClass|)$. In practice, this means that \ourmethod is fast, and only takes seconds up to a few minutes depending on the size of the data.

\section{Related Work}\label{sec:related}

Causal inference from observational data is an important open problem that has received a lot of attention in recent years~\cite{budhathoki:16:origo, mooij:16:pairs, sgouritsa:15:cure, pearl:09:book}. Traditional constraint based approaches, such as conditional independence test, require at least three random variables and can not decide between Markov equivalent causal DAGs~\cite{pearl:09:book, verma:90:markov-equiv}. In this work, we focus specifically on those cases where we have to decide between the Markov equivalent DAGs $\XtoY$ and $\YtoX$.

A well studied framework to infer the causal direction for the two-variable case relies on the additive noise assumption~\cite{shimizu:06:anm}. Simply put, it makes the strong assumption that $Y$ is generated as a function of $X$ plus additive noise, $Y = f(X) + N$, with $X \independent N$. It can then be shown that while such a function is admissible in the causal direction, this is not possible in the anti-causal direction. There exist many approaches based on this framework that try to exploit linear~\cite{shimizu:06:anm} or non-linear functions~\cite{hoyer:09:nonlinear} and can be applied to real valued~\cite{shimizu:06:anm,hoyer:09:nonlinear,zhang:09:ipcm,peters:14:continuousanm} as well as discrete data~\cite{peters:10:discreteanm}. Recently, Mooij et al.~\cite{mooij:16:pairs} reviewed several ANM-based approaches from which ANM-pHSIC, a method employing the Hilbert-Schmidt Independence Criterion (HSIC) to test for independence, performed best. 
For ANMs the confidence value is often expressed as the negative logarithm of the p-value from the used independence test~\cite{mooij:16:pairs}. P-values are, however, quite sensitive to the data size~\cite{anderson:00:p-values}, which leads to a less reliable confidence value. As we will show in the experiments, our score is robust and nearly unaffected by the data size. 

Another class of methods rely on the postulate that if $\XtoY$ the marginal distribution of the cause $P(X)$ and the conditional distribution of the effect given the cause $P(Y \mid X)$ are independent of each other. The same does not hold for the opposite direction~\cite{janzing:10:algomarkov}. The authors of \igci define this independence via orthogonality in the information space. Practically, they define their score using the entropies of $X$ and $Y$~\cite{janzing:12:igci}. Liu and Chan implemented this framework by calculating the distance correlation for discrete data between $P(X)$ and $P(Y \mid X)$~\cite{liu:16:dc}. A third approach based on this postulate is \cure~\cite{sgouritsa:15:cure}. Here, the main idea is to estimate the conditional using unsupervised inverse Gaussian process regression on the corresponding marginal and compare the result to the supervised estimation. If the supervised and unsupervised estimation for $P(X \mid Y)$ deviate less than those for $P(Y \mid X)$, an independence of $P(X \mid Y)$ and $P(X)$ is assumed and causal direction $\XtoY$ is inferred. Although well formulated in theory, the proposed framework is only solvable for data of up to $200$ data points and otherwise relies strongly on finding a good sample of the data. 

Recently, Janzing and Sch{\"{o}}lkopf postulated that if $\XtoY$, the complexity of the description of the joint distribution in terms of Kolmogorov complexity, $K(P(X,Y))$, will be shorter when first describing the distribution of the cause $K(P(X))$ and than describing the distribution of the effect given the cause $K(P(Y \mid X))$ than vice versa~\cite{janzing:10:algomarkov, lemeire:06:causalmdl}. To the best of our knowledge, Mooij et al.~\cite{mooij:10:mml} were the first to propose a practical instantiation of this framework based on the Minimum Message Length principle (MML)~\cite{wallace:68:mml} using Bayesian priors. Vreeken~\cite{vreeken:15:ergo} proposed to approximate the Kolmogorov complexity for numeric data using the cumulative residual entropy, and gave an instantiation for multivariate continuous-valued data. Perhaps most related to \ourmethod is \textsc{Origo}~\cite{budhathoki:16:origo}, which uses MDL to infer causal direction on binary data, whereas we focus on univariate numeric data. 

\section{Experiments}
\label{sec:exps}

In this section we empirically evaluate \ourmethod. In particular, we consider synthetic data, a benchmark data set, and a real-world case study. We implemented \ourmethod in $R$ and make both the code, the data generators, and real world data publicly available for research purposes.\!\footnote{\codeurl} We compare \ourmethod to the state of the art for univariate causal inference. These include \cure~\cite{sgouritsa:15:cure}, \igci~\cite{janzing:12:igci} and \resit~\cite{peters:14:continuousanm}. From the class of ANM-based methods we compare to ANM-pHSIC~\cite{hoyer:09:nonlinear,mooij:16:pairs}, which a recent survey identified as the most reliable ANM inference method~\cite{mooij:16:pairs}. We use the implementations by the authors, sticking to the recommended parameter settings.

To run \ourmethod, we have to define the parameter $t$, which is used to normalise the data $X_i$ within a local component, on which the data $Y_i$ is fitted. Generally, the exact value of $t$ is not important for the algorithm, since it only defines the domain of the data points $X_i$, which can be compensated by the parameters of the fitted function. In our experiments, we use $t=5$ and set the precision $p$ for the parameters to three.

\subsection{Synthetic Data}

We first consider data with known ground truth. To generate such data, we follow the standard scheme of Hoyer et al.~\cite{hoyer:09:nonlinear}. That is, we first generate $X$ randomly according to a given distribution, and then generate $Y$ as $Y = f(X) + N$, where $f$ is a function that can be linear, cubic or reciprocal, and $N$ is the noise term, which can either be additive or non-additive.

\subsubsection*{Accuracy}

First, we evaluate the performance of \ourmethod under different distributions. Following the scheme above, we generate $X$ randomly from either
\begin{enumerate}
	\item a uniform distribution with $\min=-t$ and $\max=t$, where $t \thicksim \text{unif}(1,10)$,
	\item a sub-Gaussian distribution by sampling data with $\mathcal{N}(0,s)$, where $s \thicksim \text{unif}(1,10)$ and taking each value to the power of $0.7$ maintaining its sign,\!\footnote{We consider sub-Gaussian distributions since linear functions with both $X$ and $N$ Gaussian distributed are not identifiable by ANMs~\cite{hoyer:09:nonlinear}.}
	\item a binomial distribution with $p \thicksim \text{unif}(0.1,0.9)$ and the number of trials $t \thicksim \left \lceil \text{unif}(1,10) \right \rceil$, or
	\item a Poisson distribution with $\lambda \thicksim \text{unif}(1,10)$.
\end{enumerate}
Note that the binomial and Poisson distribution generate discrete data points, which with high probability results in non-deterministic pairs. To generate $Y$ we first apply either a linear, cubic or reciprocal function on $X$, with fixed parameters, and add either additive noise using a uniform or Gaussian distribution with $t, s \thicksim \text{unif}(1,\max(x) / 2)$ or non-additive noise with $\mathcal{N}(0,1)|\sin(2 \pi \nu X)| + \mathcal{N}(0,1)|\sin(2 \pi (10 \nu) X)|/4$ according to~\cite{sgouritsa:15:cure}, where we choose $\nu \thicksim \text{unif}(0.25, 1.1)$. For every combination we generate $100$ data sets of $1000$ samples each.

\begin{figure}[t]
	\includegraphics[]{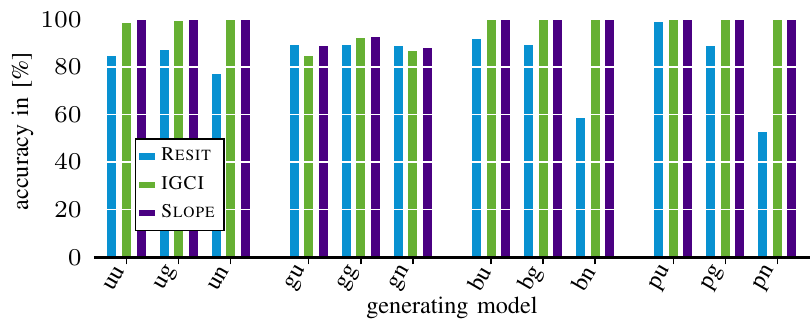}
	\caption{[Higher is better] Accuracies of \ourmethod, \resit and \igci on synthetic data. The first letter of the labels corresponds to the distribution of $X$ ($u$: uniform, $g$: sub-Gaussian, $b$: binomial and $p$: Poisson), the second letter to that of the noise ($u$: uniform, $g$: Gaussian and $n$: non-additive).}
	\label{fig:synthetic_all}
\end{figure}

Next, we apply \ourmethod, \resit, and \igci and record how many pairs they correctly infer. As they take up to hours to process a single pair, we do not consider \cure and \anm here. We give the averaged results over all three function types in Fig.~\ref{fig:synthetic_all}. In general, we find that \ourmethod and \igci perform on par and reach $100\%$ for most setups, whereas \ourmethod performs better on the sub-Gaussian data. If we consider the single results for linear, cubic and reciprocal, we find that on the linear data with sub-Gaussian distributed $X$, \ourmethod performs on average $7\%$ better than \igci. 

\subsubsection*{Confidence}

Second, we investigate the dependency of the \resit, \igci, and \ourmethod scores on the size of the data. In an ideal world, a confidence score is not affected by the size of the data, as this allows easy comparison and ranking of scores. 

\begin{figure}[t]
	\centering
	\begin{minipage}[t]{0.33\linewidth}
		\includegraphics[]{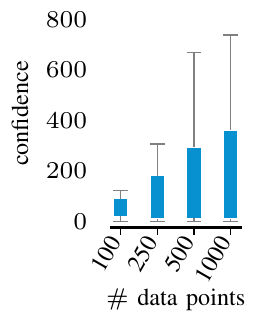}
		\subcaption{\resit}
		\label{fig:conf:resit}
	\end{minipage}%
	\begin{minipage}[t]{0.33\linewidth}
		\includegraphics[]{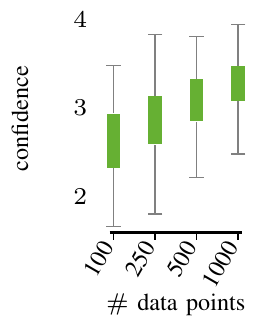}
		\subcaption{\igci}
		\label{fig:conf:igci}
	\end{minipage}%
	\begin{minipage}[t]{0.33\linewidth}
		\includegraphics[]{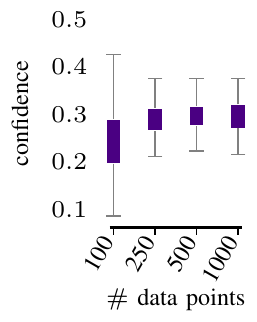}
		\subcaption{\ourmethod}
		\label{fig:conf:slope}
	\end{minipage}
	\caption{[The more stable the better] Confidence values on a cubic function for different sample sizes. Unlike \resit and \igci, the \ourmethod scores can be meaningfully compared between different sample sizes.}
	\label{fig:confidence}
\end{figure}

To analyse this, we generate $100$ datasets of $100, 250, 500$ and $1\,000$ samples each, where $X$ is Gaussian distributed and $Y$ is a cubic function of $X$ with uniform noise. Subsequently, we apply \resit, \igci and \ourmethod and record their confidence values. We show the results per sample size in Fig.~\ref{fig:confidence}. As each method uses a different score, the scale of the Y-axis is not important. What is important to note, is the trend of the scores over different sample sizes. We see  \ourmethod has very consistent confidence values that are independent of the number of samples, in strong contrast
to \resit and \igci. For \resit the standard deviation of the confidence values grows with the sample size, while for \igci, we observe that the average confidence increases with the number of samples. In other words, while it is easy to compare and rank \ourmethod scores, this is not the case for the two others---which, as we will see below results in comparatively bad decision rates.

\subsubsection*{Non-Determinacy}

Local regression on non-deterministic data adds to the modelling power of \ourmethod, yet, it may also lead to overfitting. Here we evaluate whether MDL protects us from picking up spurious structure. 

To control non-determinacy, we sample $X$ uniformly from $k$ equidistant values over $[0,1]$, i.e., $X \in [\frac{0}{k},\frac{1}{k},\cdots,\frac{k}{k}]$. To obtain $Y$, we apply a linear function and additive Gaussian noise as above. Per data set we sample $1\,000$ data points. 

\begin{figure}[t]
	\includegraphics[]{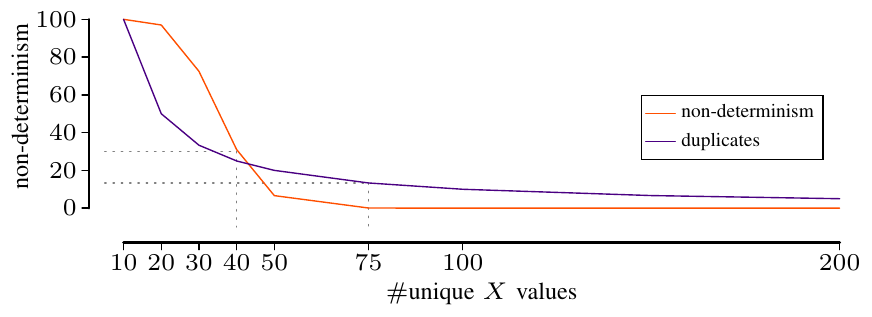}
	\caption{[\ourmethod does not overfit] Percentage of non-deterministic models \ourmethod chooses, resp. the expected number of $Y$ values per $X$ value, for the number of unique values of $X$.}
	\label{fig:overfitting}
\end{figure}

In Fig.~\ref{fig:overfitting} we plot the non-determinism of the model, i.e. the average number of used bins divided by the average number of bins \ourmethod could have used, against the number of distinct $X$ values. As a reference, we also include the average number of values of $Y$ per value of $X$. We see that for at least $75$ unique values, \ourmethod does not infer  non-deterministic models. Only at $40$ distinct values, i.e., an average of $25$ duplicates per $X$, \ourmethod consistently starts to fit non-deterministic models. This shows that if anything, rather than being prone to overfit, \ourmethod is conservative in using local models.

\subsection{Real World Data}

Next we evaluate \ourmethod on real world benchmark data. In particular, we consider the Tuebingen cause-effect data set.\!\footnote{https://webdav.tuebingen.mpg.de/cause-effect/} 
At the time of writing the data set included 98 univariate numeric cause effect pairs. 
We compare \ourmethod to \igci, \resit, \anm, and \cure, using the suggested parameter settings for this benchmark. In addition, we include \ourmethodD in our comparison, which is an ablated version of \ourmethod that only fits a single global deterministic function. 

\subsubsection*{Decision Rate and Accuracy}

We first consider the accuracy and decision rates over the benchmark data. To determine the decision rate for an approach, we order the results according to its confidence values. We then determine, from $k=1$ to $98$, the accuracy over the top-$k$ ranked pairs. To determine the accuracy at $k$ we weigh each pair according to its specification in the database. In case an algorithm does not decide, we weigh this result as one half of the corresponding weight.

\begin{figure}[t]
	\centering
	\includegraphics[]{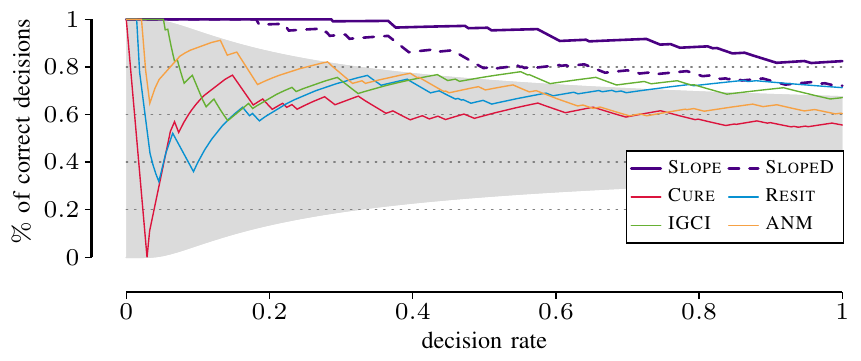}
	\caption{[Higher is better] Decision rates of \ourmethod, \cure, \resit, \igci and \anm on the Tuebingen benchmark data set (98 pairs). \ourmethodD is an ablated version of \ourmethod, which fits the data with a single deterministic function (more detailed version of Fig.~\ref{fig:decision_rate_intro}).}
	\label{fig:decision_rate}
\end{figure}

We plot the results in Fig.~\ref{fig:decision_rate}, where in addition we show the $95\%$ confidence interval for the binomial distribution with success probability of $0.5$ in gray. We observe that \ourmethod strongly outperforms its competitors in both decision rate and overall accuracy; it identifies the correct result for top-ranked $32$ data sets, over the top-$74$ pairs (which correspond to $73.8\%$ of the weights) it has an accuracy of $90\%$, while when overall it obtains an accuracy of $82.4\%$. 

\begin{figure}[t]
	\centering
   	\begin{minipage}[t]{0.95\linewidth}
	\includegraphics[]{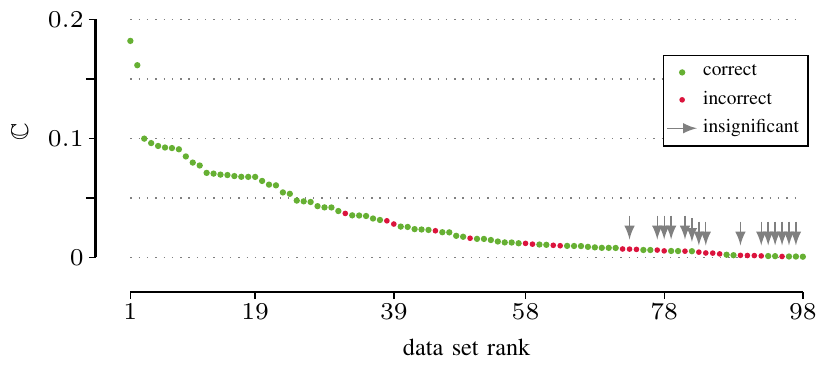}
   	\end{minipage}%
	\caption{Confidence values of \ourmethod for the Tuebingen benchmark pairs, in descending order, corresponding to Fig.~\ref{fig:decision_rate}. Correct inferences marked in green, errors in red, and inferences insignificant at $\alpha = 0.001$ marked with a gray arrow.}
	\label{fig:conf:benchm}
\end{figure}

In Fig.~\ref{fig:conf:benchm} we show the corresponding confidence values of \ourmethod for the benchmark pairs. The plot emphasises not only the predictive power of \ourmethod, but also the strong correlation between confidence value and accuracy. In comparison to the other approaches the decision rate (Fig.~\ref{fig:decision_rate}) of \ourmethod is stable and only decreases slightly at the very end. Moreover, our competitors obtain much worse overall accuracies, between 56\% (\cure) and 71\% (\resit), which for the most part are insignificant. 

We identify three reasons why \ourmethod performs better than other approaches based on functional learning. First, we consider also piecewise regression, which leads to an improvement of $10\%$ for \ourmethod compared to \ourmethodD. Second, we consider the complexity of the functions, and incorporate the model cost to prevent overfitting; as they are so powerful, methods based on Gaussian process learning, for example, are likely to overfit. Last, the calculation of our confidence value is less dependent on the data size.

If we not only consider the confidence values, but also our proposed statistical test, we can improve our results even further. After adjusting the p-values using the Benjamini-Hochberg correction~\cite{benjamini:95:fdr} to control the false discovery rate (FDR), $83$ out of the $98$ decisions are significant w.r.t. $\alpha = 0.001$. As shown in Fig.~\ref{fig:conf:benchm} the pairs rated as insignificant correspond to small confidence values. In addition, from the $15$ insignificant pairs, $9$ were inferred incorrect from \ourmethod and $6$ correct. Over the significant pairs the weighted accuracy increases to $84.9\%$. 

\begin{figure}[t]
	\centering
    	\begin{minipage}[t]{.5\linewidth}
	\includegraphics[]{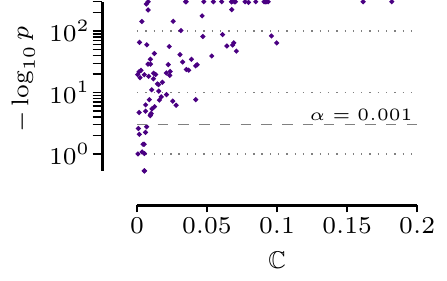}
	\subcaption{$\conf$ vs. $p$-value}
     	\label{fig:conf:pval}
   	\end{minipage}%
   	\begin{minipage}[t]{.5\linewidth}
	\includegraphics[]{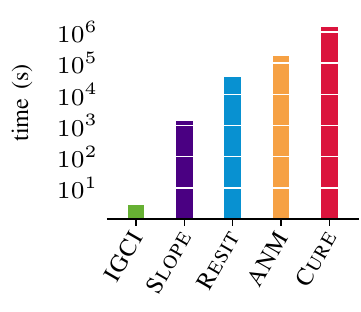}
	\subcaption{runtime}
     	\label{fig:runtime}
   	\end{minipage}%
	\caption{(left) Confidence and significance of \ourmethod on the Tuebingen benchmark pairs. Only samples with low confidence are also insignificant. (right) Runtime in seconds over all 98 pairs, in log-scale. \ourmethod is more accurate than all, and faster than all except for \igci.}
	\label{fig:runtimes}
\end{figure}

To provide further evidence that the confidence values and the p-values are indeed related, we plot the adjusted p-values and confidence values in Fig.~\ref{fig:conf:pval}. We observe that high confidence values correspond to highly significant p-values. We also computed the decision rate for \ourmethod when ranking by p-values, and find it is only slightly worse than that ranked by confidence. We posit that confidence works better as it is more independent of the data size. To test this, we calculate the correlation between data size and corresponding measures using the maximal information coefficient (MIC)~\cite{reshef:11:mic}. We find a medium correlation between confidence and p-values ($0.64$), and between p-values and data size ($0.55$), and only a weak correlation between confidence and data size ($0.31$).

To show that the strength of our method comes from its ability to fit non-deterministic models, we also consider \ourmethodD, an ablated version of \ourmethod that only considers deterministic models, i.e., only considers models that consist of a single global regression function. As it uses the same score, we see in Fig.~\ref{fig:decision_rate} that its decision rate is still good. Although the overall accuracy drops by $10\%$ to $72.1\%$, it is still  as good as the best competing method. This also shows that the full approach is much better able to deal with a broader spectrum of cause effect pairs.

Apart from the accuracies, we also tracked which functional dependencies \ourmethod found on the benchmark data. We found that most of the time ($54.6\%$), it fits linear functions. For $23.7\%$ of the data it fits exponential models, and for $15.5\%$ cubic models. Quadratic and reciprocal models are rarely fitted ($6.2\%$). This shows that working with a richer class of functions allows us to pick up more structure. In addition, we observe that although we allow to fit complex models, in many cases a simple model is preferred since it has sufficient explanatory power at lower model costs.

\subsection{Case Study: Octet Binary Semi Conductors}

To evaluate real-world performance we conduct a case study on octet binary semi-conductors~\cite{ghiringhelli:15:octet, vechten:69:quantum}. In essence, these are 82 different materials for which we have data on the radii of electron orbitals, and important other energy quantities that together determine their chemical properties. The target of interest is the energy difference $\delta_E$ between rocksalt and zincblende crystal structures. It is still an open challenge to find that combination of physical properties that can fully explain $\delta_E$~\cite{ghiringhelli:15:octet, goldsmith:17:uncovering}. For causal inference, it suffices to know that the energy difference is influenced by the descriptor variables and not the other way round.

As there is no known combination of physical properties that fully describes $\delta_E$, we mine the top $10$ features that each explain as much of $\delta_E$ as possible~\cite{mandros:17:noname}, and therewith obtain $10$ cause effect pairs where we set $\delta_E$ as $X$ and one of the mined features as $Y$. After consulting with domain experts, we assume $\YtoX$ as ground truth for all pairs. 

We find that \ourmethod infers the correct direction for $9$ out of $10$ pairs. The only error is also the only insignificant score ($p=0.199$) at $\alpha = 0.001$. In comparison, we find that \cure infers all pairs correctly, whereas \igci makes the same decisions as \ourmethod. \resit and \anm, on the other hand, only get $4$ resp. $5$ pairs correct.

\subsection{Runtime}

Last, we evaluate the computational efficiency of \ourmethod. To this end we report, per method, the wall-clock time needed to decide on all 98 pairs of the benchmark data set. We ran these experiments on Linux servers with two six-core Intel Xenon E5-2643v2 processors and 64GB RAM. The implementations of \ourmethod, \igci and \resit are single-threaded, whereas \anm and \cure are implemented in Matlab and use the default number of threads. We give the results in Fig.~\ref{fig:runtimes}. We see that \igci is fastest, followed by \ourmethod, which takes $1\,475$ seconds to processes all pairs. The other competitors are all at least one order of magnitude slower. Taking $13$ days, \cure has the longest runtime. The large gain in runtime of \ourmethod compared to \resit, \anm and \cure rises from the fact that those methods employ Gaussian process regression to fit the functions.

\section{Discussion}
\label{sec:discussion}

The experiments clearly show that \ourmethod works very well. It performs well in a wide range of settings, both on synthetic and real world data. In particular on the latter it outperforms the state of the art, obtaining highly stable decision rates and an overall accuracy of more than $10\%$ better than the state of the art. Our case study showed it makes sensible decisions. Most importantly, \ourmethod is simple and elegant. Its models are easy to interpret, it comes with a stable confidence score, a natural statistical test, and is computationally efficient. 

The core idea of \ourmethod is to decide for the causal direction by the simplest, best fitting regression function. To deal with non-deterministic data, we allow our model to additionally use local regression functions for non-deterministic values of $X$, which the experiments show leads to a large increase in performance. Importantly, we employ local regression within an MDL framework; without this, fitting local regressors would not make sense, as it would lead to strong overfitting.

A key advantage of our MDL-based instantiation of the algorithmic Markov condition, compared to HSIC-based independence tests and \igci, is that our score is not dependent on the size of the data. This makes it possible to meaningfully compare results among different tests; this is clearly reflected in the stable decision rates. Another advantage is that it allows us to define a natural statistical test based on compression rates, which allows us to avoid insignificant inferences.

Although the performance of \ourmethod is impressive, there is always room for improvement. With regard to confidence and significance, we are highly interested in investigating whether we can define a test that directly infers the significance of a confidence score (without resorting to permutation testing).

In addition, it is possible to improve the search for local components by considering alternate re-distributions of $X'$, apart from uniformly ascending values. This is not trivial, as there exist $n!$ possible orders, and it is not immediately clear how to efficiently optimise regression fit over this space. More obviously, there is room to expand the set of function classes that we use at the moment---kernel based, or Gaussian-process based regression are powerful methods that, at the expense of computation, will likely improve performance further.

For future work, we additionally aim to consider the detection of confounding variables---an open problem that we believe our information theoretic framework naturally lends itself to---as well as to extend \ourmethod to multivariate and possibly mixed-type data. We are perhaps most enthusiastic about leveraging the high accuracy of \ourmethod towards inferring causal networks from biological processes without the need of conditional independence tests. 

\section{Conclusion}\label{sec:conclusion}

We studied the problem of inferring the causal direction between two univariate numeric random variables $X$ and $Y$. To model the causal dependencies we proposed an MDL-based framework employing local and global regression. Further, we proposed \ourmethod, an efficient linear-time algorithm, to instantiate this framework. In addition, we introduced $10$ new cause effect pairs from a material science data set.

Empirical evaluations on synthetic and real world data show that \ourmethod reliably infers the correct causal direction with an high accuracy. On benchmark data, at over $82\%$ accuracy \ourmethod outperforms the state of the art by more than $10\%$, provides a more robust decision rate, while additionally also being computationally more efficient. In future research, we plan to refine our statistical test, consider detecting confounding, causal inference on multivariate setting, and use \ourmethod to infer causal networks directly from data.

\section*{Acknowledgment}
The authors wish to thank Panagiotis Mandros and Mario Boley for help with the octet binary causal pairs. Alexander Marx is supported by the International Max Planck Research School for Computer Science (IMPRS-CS). Both authors are supported by the Cluster of Excellence Multimodal Computing and Interaction within the Excellence Initiative of the German Federal Government.

\balance
\bibliographystyle{IEEEtranS}
\bibliography{abbrev,bib-jilles,bib-paper,bib-alex}


\end{document}